\documentclass[letterpaper, 10 pt, conference]{ieeeconf}  

\IEEEoverridecommandlockouts                              

\usepackage[dvipsnames]{xcolor}

\usepackage{pgfplots}
\pgfplotsset{compat=newest}
\usetikzlibrary{plotmarks}
\usetikzlibrary{arrows.meta}
\usepgfplotslibrary{patchplots}
\usepackage{grffile}
\usepackage{amsmath}
\usepackage{blindtext}
\usepackage[bottom]{footmisc}
\usepackage{mathrsfs}
\usepackage{lipsum}
\usepackage{url}
\usepackage{array}
\usepackage{booktabs}
\usepackage{float}
\usepackage{multirow}
\usepackage{cite}
\newlength\fwidth
\usepackage{graphicx}
\usepackage{framed}
\usepackage{indentfirst}
\usepackage{makecell}

\usepackage{amsmath, amssymb,amsthm}
\usepackage{arydshln}
\usepackage{color}
\usepackage{tabularx}
\usepackage{hyperref}
\usepackage[capitalise]{cleveref}
\usepackage{comment}
\usepackage{romannum}
\usepackage{subfig}
\usepackage[]{algorithm, algpseudocode}
\usepackage{balance}

\usepackage{dsfont}
\usepackage{cancel}

\usepackage{colortbl}

\usepackage{tabu}
\usepackage{xcolor}
\usepackage{wrapfig}

\definecolor{mygray1}{gray}{.9}
\definecolor{mygray2}{gray}{.7}
\usepackage{bm}
\usepackage{bbm}
\definecolor{mygray}{gray}{.92}
\makeatletter
\newcommand{\thickhline}{%
    \noalign {\ifnum 0=`}\fi \hrule height 1pt
    \futurelet \reserved@a \@xhline
}

\usepackage{tikz}

\title{\LARGE \bf
Cross-Modal Self-Supervised Learning with Effective Contrastive Units for LiDAR Point Clouds
}

\author{Mu Cai$^{1}$ \quad Chenxu Luo$^{2}$ \quad Yong Jae Lee$^{1}$ \quad Xiaodong Yang$^{2}$ 
\thanks{$^{1}$Mu Cai and Yong Jae Lee are with the Department of Computer Sciences, University of Wisconsin-Madison, Madison, WI 53706, USA. {\tt\small \{mucai,yongjaelee\}@cs.wisc.edu} This work was conducted during Mu's internship at QCraft.}%
\thanks{$^{2}$Chenxu Luo and Xiaodong Yang are with QCraft, Santa Clara, CA 95054, USA. {\tt\small \{chenxu,xiaodong\}@qcraft.ai}
   }%
\thanks{This work was supported in part by the Institute of Information \& Communications Technology Planning \& Evaluation (IITP) grant funded by the Korean government (MSIT) (No. 2022-0-00871, Development of AI Autonomy and Knowledge Enhancement for AI Agent Collaboration).}
}

\begin{document}

\maketitle
\thispagestyle{empty}
\pagestyle{empty}



\begin{abstract}
3D perception in LiDAR point clouds is crucial for a self-driving vehicle to properly act in 3D environment. However, manually labeling point clouds is hard and costly. There has been a growing interest in self-supervised pre-training of 3D perception models. Following the success of contrastive learning in images, current methods mostly conduct contrastive pre-training on point clouds only. Yet an autonomous driving vehicle is typically supplied with multiple sensors including cameras and LiDAR. In this context, we systematically study single modality, cross-modality, and multi-modality for contrastive learning of point clouds, and show that cross-modality wins over other alternatives. In addition, considering the huge difference between the training sources in 2D images and 3D point clouds, it remains unclear how to design more effective contrastive units for LiDAR. We therefore propose the instance-aware and similarity-balanced contrastive units that are tailored for self-driving point clouds. Extensive experiments reveal that our approach achieves remarkable performance gains over various point cloud models across the downstream perception tasks of LiDAR based 3D object detection and 3D semantic segmentation on the four popular benchmarks including Waymo Open Dataset, nuScenes, SemanticKITTI and ONCE. 
\end{abstract}    
\section{Introduction}
\label{sec:intro}

\begin{figure}[h]
  \centering
  \includegraphics[width=\linewidth]{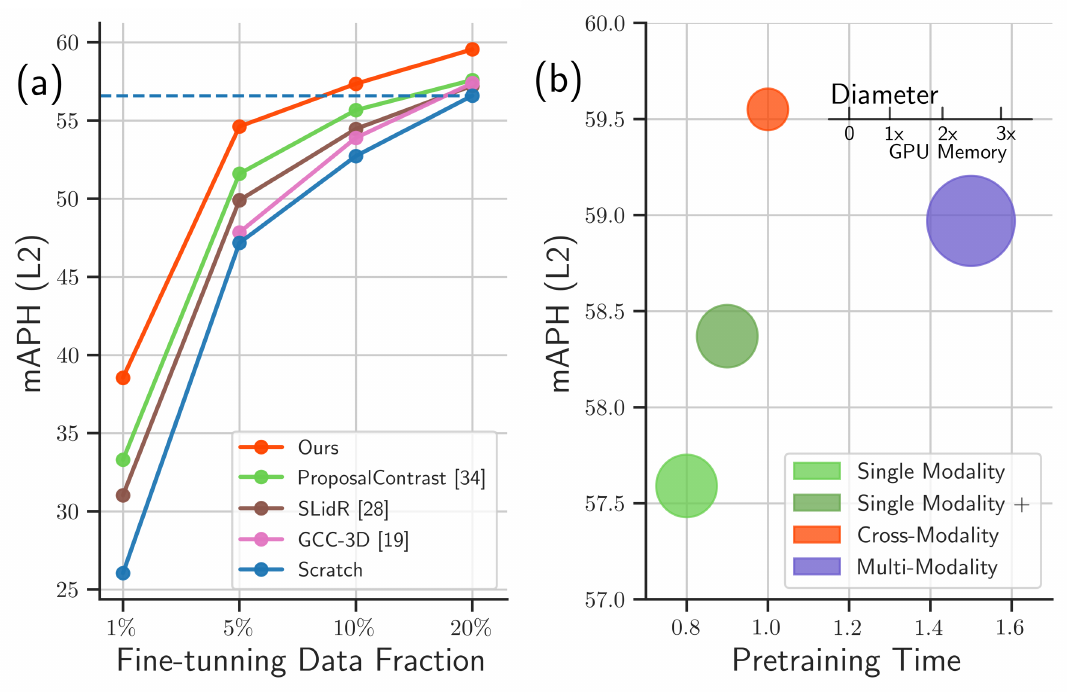}
  \vspace{-4mm}
   \caption{(a) Our approach achieves consistent and significant performance gains compared to training from scratch and other state-of-the-art self-supervised learning methods for LiDAR point clouds across different fractions of fine-tuning data on Waymo Open Dataset. (b) Our comprehensive modality study finds that cross-modality (ours) is superior to single modality (and its enhanced version +) and multi-modality in terms of downstream performance and memory consumption of GPU (proportional to bubble area), while requiring moderate pre-training time. } 
   \vspace{-4mm}
   \label{fig:teaser_paper}
\end{figure}

3D perception is a pivotal module for an autonomous driving vehicle as it provides the fundamental information to subsequent onboard modules ranging from prediction to planning~\cite{wang2023prophnet, liu2023rlil, li2023tip}. LiDAR is one of the most commonly utilized sensor that a self-driving system relies on to perceive its neighboring environment in 3D~\cite{li2023pillarnext}. However, annotating LiDAR point clouds is notoriously difficult, error-prone, and time-consuming. 
For instance, it costs around 4.5 hours to label a single tile in SemanticKITTI~\cite{behley2019semantickitti}. Recently, there has been growing attention in making use of self-supervised learning (SSL) to alleviate the laborious human labeling efforts, and at the same time, to harvest the vast amount of data continuously collected by the world-wide self-driving fleets. However, 3D SSL is still under explored compared to the well-developed family of 2D SSL methods~\cite{byol, beit, chen2021empirical, pmlr-v119-chen20j}.

As the pioneers in 3D SSL,  DepthContrast~\cite{Zhang_2021_ICCV} conducts contrastive pre-training by using the holistic point cloud as a contrastive unit at the scene-level, while PointContrast~\cite{xie2020pointcontrast} performs point-level comparisons in two transformed point clouds with different views to capture dense
information at the point-level. Such methods are designed for indoor settings captured by hundreds of scans from diverse positions per scene with limited occlusion. 
In contrast, LiDAR point clouds in autonomous driving capture large-scale outdoor scenes with restricted viewing angles and strong occlusions. Most LiDAR point clouds are very similar to each other from the scene-level perspective as a result of the limited diversity in street views. These differences make such scene-level 3D SSL methods incompatible with self-driving point clouds.

Recently, GCC-3D~\cite{Liang_2021_ICCV} and ProposalContrast~\cite{yin2022proposalcontrast} propose to generate more fine-grained contrastive units in the region-level for LiDAR. They leverage preliminary geometric cues 
to drive contrastive pre-training. 
However, our experiments reveal that using low-level geometry makes the self-supervised objective easy to overfit and leads to the sub-optimal performance in downstream tasks.

Another track is to perform contrastive learning across images and point clouds. Pri3D~\cite{hou2021pri3d} and PPKT~\cite{liu2021ppkt} take the first step in exploring pixel-point correspondence 
for indoor point clouds. 
SLidR~\cite{SLidR} uses LiDAR point clouds and synchronized images to carry out contrastive learning, where superpixels are used to group local pixels as contrastive units. However, superpixels tend to over-segment an object into small fragments, leading to numerous false negative pairs and imbalanced sampling in the contrastive objective.
Our experiments show that the pre-trained weights provided by SLidR deliver on par or even deteriorated results compared to the randomly initialized weights when fine-tuning on the downstream (large-scale annotated) datasets.

\begin{figure}[t]
  \centering
  \includegraphics[width=\linewidth]{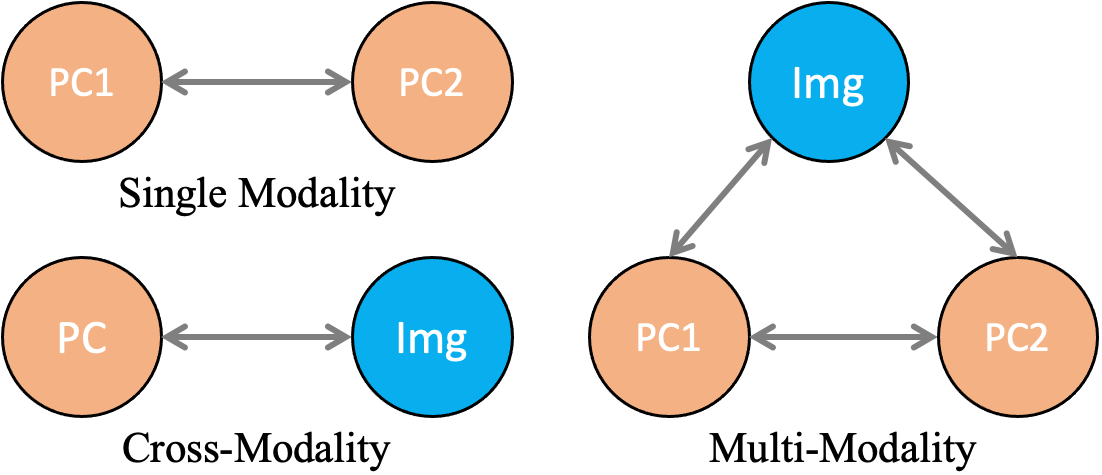}
  \vspace{-4mm}
   \caption{Illustration of the single modality, cross-modality, and multi-modality for contrastive learning of LiDAR point clouds. PC1 and PC2 denote two independently augmented point clouds.} 
   \label{fig:modality}
   \vspace{-5mm}
\end{figure}

\begin{figure*}[t]
  \centering
  \includegraphics[width=0.98\linewidth]{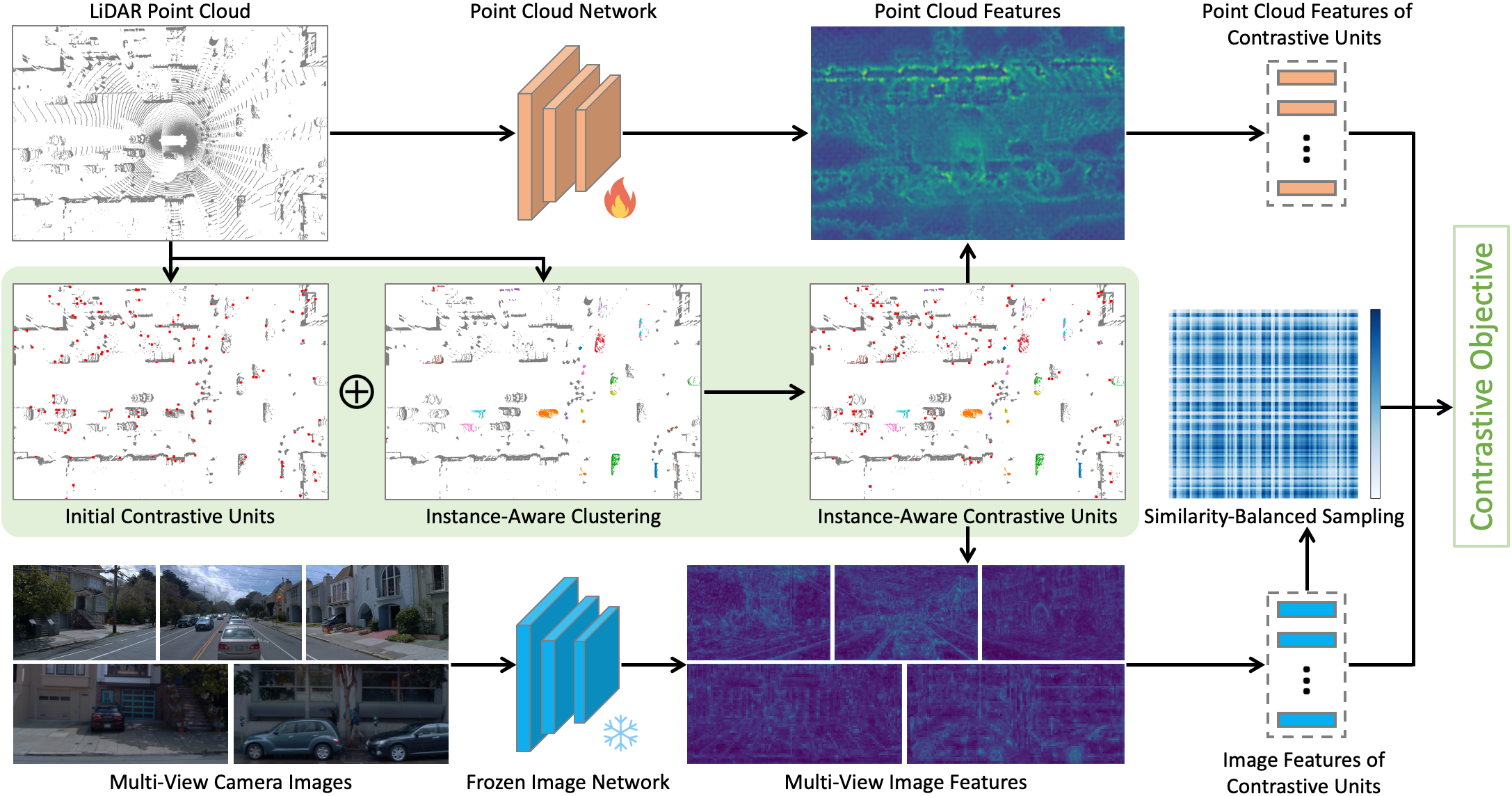} 
   \caption{Overview of the proposed cross-modal contrastive pre-training framework. We uniformly sample initial contrastive units to maximally cover the point cloud scene. An unsupervised geometry clustering is introduced to generate the instance-aware contrastive units. Leveraging on the image features that are self-supervised pre-trained with rich semantics, we develop the similarity-balanced sampling to balance the contrastive objective by ruling out those units that are semantically close.}
   \vspace{-5mm}
   \label{fig:teaser}
\end{figure*}

In light of the above observations, we seek to answer the two fundamental research questions for LiDAR based 3D SSL: (1) \textit{which modalities are better suited for contrastive learning of point clouds}, and (2) \textit{how to design more effective contrastive units in self-driving scenarios}. 

First, an autonomous vehicle is typically equipped with a sensor suite including cameras and LiDAR~\cite{wang2023distill}, offering three possible modalities to perform contrastive pre-training on: (i) single modality with point clouds only, (ii) cross-modality on images and point clouds, and (iii) multi-modality by combining (i) and (ii), as depicted in Figure~\ref{fig:modality}. We find that \textit{the cross-modality wins over the other two alternatives in terms of both pre-training efficiency and downstream improvement}, as shown in Figure~\ref{fig:teaser_paper}. Specifically, we show that the contrastive learning on point clouds only is prone to overfit to the pre-training objective, while the multi-modality induces tremendous extra memory and computational costs yet brings no additional performance gains.

Second, a huge discrepancy exists in the training sources between 2D and 3D SSL. ImageNet~\cite{imagenet_cvpr09} is the de facto training data for 2D SSL, and it is essentially a curated dataset that is instance-concentrated and class-balanced. On the contrary, real-world driving data is naturally collected at the scene-level consisting of an imbalanced compound of multiple instances and vast background with no specific focus. Inspired by this contrast, we devise \textit{the instance-aware and similarity-balanced contrastive units} in 3D SSL to approximate the counterpart in 2D SSL. In practice, we sample the initial contrastive units uniformly in a point cloud to ensure a thorough coverage of the scene. An unsupervised geometry clustering method is then introduced to merge and grow a part of the initial units into instance clusters to create the instance-aware contrastive units. As demonstrated in Figure~\ref{fig:cluster_vis}(a), we can discover a rich set of foreground instances such as vehicles and pedestrians via the clustering. For the remaining initial units, a large portion are similar and monotonous from the wide-open background such as vegetation and buildings. To better balance the contrastive objective, we develop a similarity-balanced sampling to rule out the semantically similar units.   

Our main contributions are summarized as follows. (1) To our knowledge, this work provides the first comprehensive study in term of modality for contrastive learning of point clouds in self-driving scenarios. Our findings demonstrate that the cross-modal learning across images and point clouds performs the best in pre-training efficiency and accuracy improvement for downstream tasks. (2) We propose the instance-aware and similarity-balanced contrastive units such that contrastive pre-training can be conducted at the instance-level with more balanced sampling. (3) Experiments reveal that our approach achieves superior performance gains on multiple downstream tasks, as shown in our extensive evaluations. 
For instance, our pre-trained weights boost the training-from-scratch performance by 2.96\% L2 mAPH on Waymo Open Dataset, exceeding the previous best result~\cite{yin2022proposalcontrast} by 1.91\%. Our code and models will be released at \url{https://github.com/qcraftai/cross-modal-ssl}.

\section{Method}
In this section, we detail the proposed instance-aware and similarity-balanced contrastive units in cross-modal 3D SSL. We start by introducing the point cloud and image feature representations, and then describe the design of our contrastive units including instance-aware clustering and similarity-balanced sampling. Finally, we present our contrastive pre-training objective.   

\subsection{Feature Representations} 
\label{sec:crossmodality}

\noindent\textbf{Point Cloud Feature Representation.} 
Data augmentation is important to 
contrastive learning since it increases the difficulty of self-supervised learning, alleviates overfitting, and encourages the pre-trained weights to learn invariant features. Given a point cloud $\mathcal{P}\in \mathbb{R}^{N \times 3}$ with $N$ points, we first apply a set of transformations $\mathcal{T}$ to $\mathcal{P}$, resulting in the augmented point cloud $\mathcal{T(P)}$. In this paper, we use rotation, scaling, and random flipping as the augmentation set. 

We denote $F_{\text{point}}$ as a point cloud network to be self-supervised pre-trained. It is used to generate the point cloud feature $P=F_{\text{point}}(\mathcal{T(P)})$, where $P \in \mathbb{R}^{N \times C}$ and $C$ is the feature dimension. The goal of pre-training is to enable $F_{\text{point}}$ to learn the high-level semantics that are essential for the downstream perception tasks, but with no data labeling.  
Our approach is versatile to various network architectures, including the point, voxel or pillar based models.

\noindent\textbf{Image Feature Representation.}  
Along with LiDAR point clouds, the synchronized multi-view images in a self-driving vehicle provide extra visual information. Built upon the large-scale and well-established image datasets such as ImageNet~\cite{imagenet_cvpr09}, current self-supervised pre-trained networks such as MoCo~\cite{chen2021empirical} and SimCLR~\cite{pmlr-v119-chen20j} provide high-quality image features with rich semantics. We therefore take advantage of such a frozen pre-trained network as the image encoder, which brings the following three benefits. First, leveraging on the success of 2D SSL, the image features learned with high-level semantics can guide the contrastive pre-training toward the high-level understanding beyond the low-level point cloud statistics. Second, the image features involving visual texture and context offer complementary information in addition to the geometric cues from point clouds. Third, the image features are frozen and utilized as ``anchors'' to prevent contrastive learning from overfitting. As demonstrated in our experiments, only using the point cloud features tends to lead to a ``shortcut'' of geometry to fulfill the pre-training objective and lack of the desired understanding in semantics.

For a point cloud $\mathcal{P}$, which is paired with $M$ synchronized images $\{\mathcal{I}_i \in \mathbb{R}^{H \times W \times 3}, i=1,\dots, M\}$, we use a frozen pre-trained image network $F_{\text{image}}$ to generate each image feature as $I_i = F_{\text{image}}(\mathcal{I}_i)$, where $I_i \in \mathbb{R}^{H’ \times W’ \times C'}$ and $ H’, W’, C'$ denote the feature map dimensions. In our implementation, we adopt multi-scale image features from different abstraction levels as the final representation, by upsampling and concatenating feature maps from multiple resolutions. This is found to be beneficial for the downstream 3D detection and segmentation tasks in point clouds.

\noindent\textbf{Correspondence.}
It is straightforward to set up the correspondence between image features and point cloud features in the self-driving data. With the available calibration parameters between cameras and LiDAR, we project 3D point coordinates into 2D pixel coordinates to form the correspondence. When sampling points for initial contrastive units, we only consider the points that can be projected into at least one camera canvas. As for those points that can be projected into multiple cameras, we simply use average pooling of their corresponding features to obtain the final image representation.

\subsection{Contrastive Units}
\label{sec:Contrastive Unit Design}

\noindent\textbf{Ground Removal.}
For LiDAR point clouds captured in autonomous driving, a great deal of points are collected on the ground. Sampling such points results in uninformative contrastive units in the pre-training objective, hindering the learning of true foreground objects that are more relevant for the downstream tasks. Thus, we apply a simple unsupervised ground segmentation algorithm~\cite{himmelsbach2010fast} to identify and remove the LiDAR returns from ground. As shown in Figure~\ref{fig:teaser}, ground removal provides a more effective sampling space to generate contrastive units.

\noindent\textbf{Initial Contrastive Units.} 
We start from sampling individual points in the ground-removed point cloud as the initial contrastive units. Due to the scanning mechanism of LiDAR, point clouds are extremely uneven, i.e., the point density close to the ego-vehicle is tremendously higher than that far away. To initially acquire a thorough coverage of the entire scene, we utilize the farthest point sampling (FPS). In addition, inspired by recent perception works in bird's eye view (BEV)~\cite{luo2021pillar}, we ignore the height dimension in FPS. To supply an initial unit with more context, we further sample and aggregate the features from its neighboring points, which can be sampled by either $K$ nearest neighbor points or all points inside the pillar centered at the initial unit. Given the point cloud features or image features of an initial unit and its contextual points, we apply average pooling to get the corresponding representations of the unit.   
\label{sec:initial}

\begin{figure}[t]
  \centering
  \includegraphics[width=\linewidth]{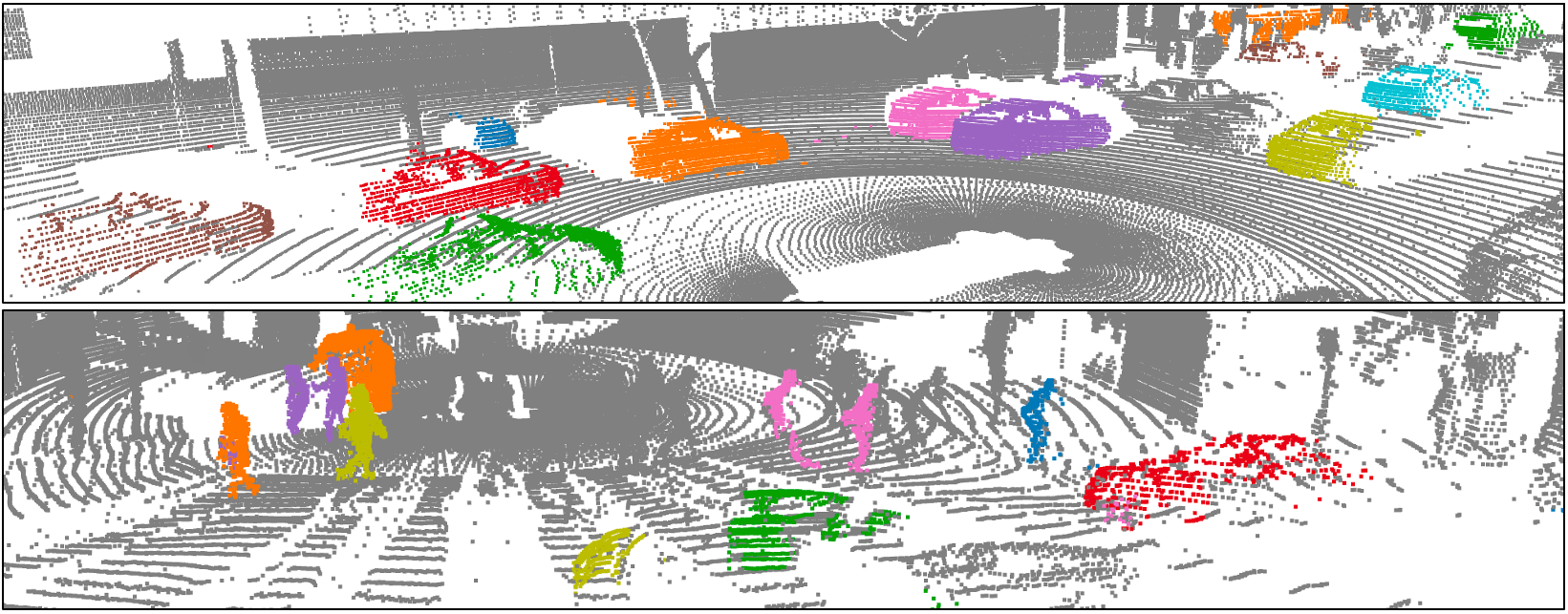}
   \caption{Illustration of instances such as vehicles and pedestrians discovered by the unsupervised clustering. 
   Note that some instances are missing due to the imperfection of the simple rule based clustering. } 
   \label{fig:cluster_vis}
   \vspace{-5mm}
\end{figure}

\noindent\textbf{Instance-Aware Clustering.} 
Though the initial contrastive units are designed to maximally cover a scene, one foreground instance such as a vehicle can be segmented into several different units, as can be seen in Figure~\ref{fig:teaser}. This results in undesirable negative pairs in the contrastive objective and is detrimental to the learning of semantics at the object or instance level. Fortunately, unlike images, point clouds possess accurate geometric measurements, making it possible to discover instances in an unsupervised manner. Here we use a simple geometry clustering algorithm~\cite{klasing2008clustering} after sampling the initial contrastive units using FPS, which employs a k-d tree to cluster all neighboring points within a radius as one instance in the range image. Here k-d tree is used to gradually refine the discovered clusters, and the range image is a 2D representation of the point cloud from range view~\cite{range}. We then filter out the clusters with anomalous sizes or aspect ratios. As illustrated in Figure~\ref{fig:cluster_vis}, we discover plenty of clusters or instances with meaningful semantics.

For the initial contrastive units that do not fall into any of the filtered clusters, their feature representations remain the same. As for the ones that fall into the same cluster, we merge them into a single unit, and then apply average pooling on their corresponding features to obtain the instance-level representation of the merged unit. In this way, we are able to substantially reduce the false negative pairs that are initially sampled from same instances, and meanwhile, to promote the contrastive units from initial points with relatively limited neighboring context to be instance-aware.

\noindent\textbf{Similarity-Balanced Sampling.}
A lightweight multi-layer perceptron is applied as the projection head to map the image features $\{I_i\}$ and point cloud features $\{P_i\}$ of the instance-aware contrastive units to the final representations $\{\tilde{I}_i\}$ and $\{\tilde{P}_i\}$ to compute the contrastive objective. A straightforward way to conduct the contrastive learning is to exploit the corresponding cross-modal features from the same unit as a positive pair and all different units as negative pairs. However, due to the extreme foreground-background imbalance in LiDAR point clouds, numerous semantically similar units can be treated as negative pairs in the contrastive objective. For instance, if a unit is sampled from the vegetation, then other vegetation units with similar semantics would constitute a great portion of the negative pairs, as shown in Figure~\ref{fig:teaser}, misleading the contrastive pre-training. This is an inherent difficulty for LiDAR as self-driving point clouds are dominated by vast background.    

By leveraging the frozen image network pre-trained with rich semantics, we can take advantage of the similarity of image features to reflect how semantically close any two contrastive units are. Given two units in a point cloud, we denote their corresponding image features as $\tilde{I}_i$ and $\tilde{I}_j$, and use the cosine similarity $s_{i j}=\left\langle \frac{\tilde{I}_i}{||\tilde{I}_i||_2} , \frac{\tilde{I}_j}{||\tilde{I}_j||_2}\right\rangle$ to measure their semantic similarity, as visualized in Figure~\ref{fig:teaser}. Based on this measurement, we propose the similarity-balanced negative sampling strategy. Given the $i$-th unit, we measure its similarity to all other units, and keep the least similar $L$ units to form the negative pairs involved in the contrastive objective. Let the $L$-th least similarity be $s_{iL}$, then the similarity-balanced negative set for the $i$-th unit is $\mathcal{S}_i = \{j \mid s_{ij} < s_{iL}, j = 1,...,B, j \neq i\}$, where $B$ is the number of instance-aware contrastive units. 
After excluding the negative pairs with high similarity in semantics, we obtain a more expressive negative set for each unit.

\subsection{Contrastive Objective}
\label{sec:Pretraining Objective}
Based on the instance-aware and similarity-balanced contrastive units, we compute the contrastive objective by InfoNCE~\cite{pmlr-v119-chen20j} on both image features and point cloud features of each unit for pre-training. The overall objective can be formalized as image-point cloud feature matching:

\begin{align*}
\begin{split}
    \mathcal{L}  = & -\frac{1}{2B} \sum_{i=1}^{B} \log \left[\frac{e^{\left(\left\langle\tilde{I}_i, \tilde{P}_i\right\rangle / \tau\right)}}{  \sum_{j \in \mathcal{S}_i}   e^{\left(\left\langle\tilde{I}_i, \tilde{P}_j\right\rangle / \tau\right)}+e^{\left(\left\langle\tilde{I}_i, \tilde{P}_i\right\rangle / \tau\right)}}\right] \\
    & - \frac{1}{2B} \sum_{i=1}^{B} \log \left[\frac{e^{\left(\left\langle\tilde{P}_i, \tilde{I}_i\right\rangle / \tau\right)}}{  \sum_{j \in \mathcal{S}_i}    e^{\left(\left\langle\tilde{P}_i, \tilde{I}_j\right\rangle / \tau\right)}+e^{\left(\left\langle\tilde{P}_i, \tilde{I}_i\right\rangle / \tau\right)}}\right], 
\end{split}
\end{align*}
where $\tau$ is the temperature and $B$ is the number of contrastive units after instance-aware clustering.

\section{Experiments}

We conduct extensive experiments on four datasets including Waymo Open Dataset (WOD)~\cite{sun2020waymo}, nuScenes~\cite{nuscenes}, SemanticKITTI~\cite{behley2019semantickitti}, and ONCE~\cite{mao2021one}. 
Our approach is applicable to various point cloud models. We select three representative networks in our experiments for fair comparison with previous works: CenterPoint (both pillar and voxel versions)~\cite{yin2021center} and MinkowskiNet~\cite{xie2020pointcontrast}. 
As for the image based network, we use the self-supervised pre-trained ResNet50~\cite{chen2021empirical} to extract image features.

\subsection{Comparison with State-of-the-Art Methods} 

\noindent\textbf{Comparison on WOD.} 
We first use CenterPoint-Pillar and follow the standard fine-tuning protocol using 30 epochs and 20\% of training samples of WOD. As shown in Table~\ref{table:waymo-CenterPoint-PointPillars}, our approach achieves the most significant performance gain compared to training from scratch, even though GCC-3D is built upon a lower baseline (relatively easier to produce a larger gain on a lower baseline). 
As expected, the overall performance of point-level pre-training in PointContrast is inferior due to the fact that its granularity (dense points as contrastive units) is not suited for self-driving point clouds. 
Among the cross-modal methods, our approach largely outperforms SLidR thanks to our design of contrastive units that are instance-aware and similarity-balanced.  
Our approach remarkably improves training from scratch by 2.73\% mAP and 2.96\% mAPH. In particular, we observe greater boost on pedestrians (+3.65\% APH) and cyclists (+3.03\% APH) compared to vehicles (+2.21\% APH). In contrast, ProposalContrast receives larger improvement on vehicles (+1.40\% APH) than pedestrians (+1.07\% APH) and cyclists (+0.54\% APH). This validates the advantage of visual cues provided in image features for the contrastive learning of small objects in point clouds, and pre-training on point clouds only makes it hard to guide the learning of small objects. 

\begin{table*}[t]
\small
\centering
\renewcommand\arraystretch{1.1}
\begin{tabular}{c|c|c|ccc}
\toprule
\rowcolor[HTML]{EFEFEF}  &\textbf{~Performance} & \textbf{Overall}             & \textbf{Vehicle}           & \textbf{Pedestrian}            & \textbf{Cyclist}          \\  \rowcolor[HTML]{EFEFEF} \multirow{-2}{*}{\textbf{Pre-training}}
&\textbf{~Gain} & \textbf{mAP/mAPH}             &  \textbf{AP/APH}           &  \textbf{AP/APH}            &  \textbf{AP/APH}          \\ \hline

Scratch$^*$               &  -   & 59.14/55.25   & -    & - 	   & -    \\
PointContrast$^*$~\cite{xie2020pointcontrast}               &  0.90/1.06     & 60.04/56.31   & -    & - 	   & -     \\

GCC-3D$^*$~\cite{Liang_2021_ICCV}                  &  2.44/2.14& 61.58/57.39   & -    & - 	   & -      \\
\hline
Scratch               &  -  & 60.74/56.59   & 62.03/61.46	   & 61.70/51.68	   &58.49/56.63   \\

PPKT~\cite{liu2021ppkt}      &       0.53/0.51      & 61.27/57.10  &  62.62/62.09   & 62.24/52.17  & 58.95/57.04    \\

SLidR~\cite{SLidR}                  &  0.66/0.64 & 61.40/57.23   & 62.40/61.87   & 62.49/52.20 & 59.30/57.64   \\

SegContrast~\cite{segcontrast}   &   0.54/0.49   & 61.28/57.08   & 62.44/61.90   & 62.39/52.10  & 59.00/57.24     \\

%

ProposalContrast~\cite{yin2022proposalcontrast}                 & 0.88/1.00     & 61.62/57.59 & 63.42/62.86  & 62.38/52.75 & 59.07/57.17    \\

Ours             &   \textbf{2.73/2.96}  & \textbf{63.47/59.55}  & \textbf{64.22/63.67}  & \textbf{64.69/55.33}  & \textbf{61.49/59.66}    \\ \bottomrule
\end{tabular}
\caption{Comparison of 3D object detection based on CenterPoint-Pillar. We report the results of L2 AP and APH on the validation set of WOD. ${ }^*$ denotes the results from~\cite{Liang_2021_ICCV}.}
\label{table:waymo-CenterPoint-PointPillars}
\vspace{-1mm}
\end{table*}

\begin{table*}[htbp]
\small
\centering
\renewcommand\arraystretch{1.1}
\begin{tabular}{c|c|c|ccc}
\toprule
\rowcolor[HTML]{EFEFEF}  &\textbf{~Performance} & \textbf{Overall}             & \textbf{Vehicle}           & \textbf{Pedestrian}            & \textbf{Cyclist}          \\  \rowcolor[HTML]{EFEFEF} \multirow{-2}{*}{\textbf{Pre-training}}
&\textbf{~Gain} & \textbf{mAP/mAPH}             &  \textbf{AP/APH}           &  \textbf{AP/APH}            &  \textbf{AP/APH}          \\ \hline
Scratch$^*$               &  -   &63.46/60.95  & 61.81/61.30    & 63.62/57.79 	   & 64.96/63.77   \\

GCC-3D$^*$~\cite{Liang_2021_ICCV}        &           1.83/1.84   & 65.29/62.79   & 63.97/63.47    & 64.23/58.47 	   & 67.68/66.44      \\
\hline

Scratch               &  -  & 65.42/62.98  & 63.82/63.33	   &64.85/59.22	   &67.58/66.38   \\

PPKT~\cite{liu2021ppkt}      &    1.18/1.14    & 66.59/64.12  &  63.53/63.02  &  64.74/58.84 & 67.01/65.85 \\

SLidR~\cite{SLidR}       &     0.69/0.67         & 66.11/63.65 & 64.34/63.84  & 66.10/60.45 &  67.87/66.68  \\

ProposalContrast~\cite{yin2022proposalcontrast}                 & 1.01/0.93   & 66.43/63.91 & 64.65/64.13  & 66.04/60.23 & 68.59/67.37    \\

Ours             &   \textbf{1.63/1.58}  & \textbf{67.05/64.56}  & \textbf{65.29/64.78}  & \textbf{67.28/61.50}  & \textbf{68.58/67.41}    \\ \bottomrule
\end{tabular}
\caption{Comparison of 3D object detection based on CenterPoint-Voxel. We report the results of L2 AP and APH on the validation set of WOD. ${ }^*$ denotes the results from~\cite{Liang_2021_ICCV}.}
\label{table:waymo-CenterPoint-VoxelNet}
\end{table*}

\begin{table*}[htbp]
\small
\centering
\renewcommand\arraystretch{1.1}
\begin{tabular}{c|c|c|ccc}
\toprule
\rowcolor[HTML]{EFEFEF}  &\textbf{~Performance} & \textbf{Overall}             & \textbf{Vehicle}           & \textbf{Pedestrian}            & \textbf{Cyclist}          \\  \rowcolor[HTML]{EFEFEF} \multirow{-2}{*}{\textbf{Pre-training}}
&\textbf{~Gain} & \textbf{mAP/mAPH}             &  \textbf{AP/APH}           &  \textbf{AP/APH}            &  \textbf{AP/APH}          \\ \hline
Scratch$^*$               &  -   &65.60/63.21  & 64.18/63.69    & 	 65.22/59.68  & 67.41/66.25   \\

BEV-MAE$^*$~\cite{lin2022bevmae}        &           1.32/1.24   & 66.92/64.45   & 64.78/64.29    & 66.25/60.53 	   & 69.73/68.52      \\
\hline

Scratch$^\dagger$               &  -   &64.51/61.92  & 63.16/62.65    & 	64.27/58.23  & 66.11/64.87   \\

Voxel-MAE$^\dagger$~\cite{min2022occupancymae}        &    1.35/1.31       &65.86/63.23 & 64.05/63.53    &65.78/59.62 	   & 67.76/66.53      \\

MAELi$^\dagger$~\cite{krispel2022maeli}        &   1.09/1.08         & 65.60/63.00   & 64.22/63.70    & 65.93/59.79	   & 66.66/65.52     \\

\hline

Scratch               &  -  & 65.42/62.98  & 63.82/63.33	   &64.85/59.22	   &67.58/66.38   \\

Ours             &   \textbf{1.63/1.58}  & \textbf{67.05/64.56}  & \textbf{65.29/64.78}  & \textbf{67.28/61.50}  & \textbf{68.58/67.41}    \\ \bottomrule
\end{tabular}
\caption{Comparison with the generative masked modeling based self-supervised learning methods for 3D object detection based on CenterPoint-Voxel. We report the results of L2 AP and APH on the validation set of WOD. ${ }^*$ denotes the results from~\cite{lin2022bevmae} and ${ }^\dagger$ from~\cite{min2022occupancymae, krispel2022maeli}.}
\label{table:waymo-CenterPoint-VoxelNet-generative}
\vspace{-3mm}
\end{table*}

\begin{table}[t]
\centering
\renewcommand\arraystretch{1.1}
\resizebox{0.45\textwidth}{!}{
\begin{tabular}{c|ccc}
\toprule
\rowcolor[HTML]{EFEFEF}  \textbf{~Pre-training} & \textbf{mAP}             & \textbf{NDS}           & \textbf{mAP@1}          \\ \hline
Scratch$^*$     &   49.60 &  	60.20 &  	- \\

GCC-3D$^*$~\cite{Liang_2021_ICCV}    &   50.80$_{+1.20}$ & 	60.80$_{+0.60}$  & -     \\
\hline

Scratch          &     51.34 &  	61.22 &  	24.66 \\

SLidR~\cite{SLidR}       &    50.82$_{-0.52}$ &  	61.01$_{-0.21}$ &  	25.59$_{+0.93}$  \\

Ours        &    \textbf{52.91$_{\textbf{+1.57}}$} &  	\textbf{62.65$_{\textbf{+1.43}}$} &  	\textbf{33.06$_{\textbf{+8.40}}$ } \\ \bottomrule
\end{tabular}}
\caption{Comparison of 3D object detection based on CenterPoint-Pillar on the validation set of nuScenes. We report the results of mAP, NDS, and mAP at the first epoch. ${ }^*$ denotes the results from~\cite{Liang_2021_ICCV}.} 
\vspace{-7mm}
\label{table:nuscenes-CenterPoint-PointPillars}
\end{table}

We then evaluate our approach using a stronger point cloud network CenterPoint-Voxel. As shown in Table~\ref{table:waymo-CenterPoint-VoxelNet}, we find a similar trend in comparison with other contrastive learning methods, and the proposed approach still achieves superior performance with a stronger baseline or backbone.  
Furthermore, we compare with the leading generative masked modeling based 3D SSL in Table~\ref{table:waymo-CenterPoint-VoxelNet-generative}. Our approach also compares favorably with the methods in this field.

\noindent\textbf{Comparison on nuScenes.} 
We pre-train CenterPoint-Pillar on nuScenes, and then fine-tune for 20 epochs using 100\% labeled data under the strong setting of using 10 sweeps as input. Table~\ref{table:nuscenes-CenterPoint-PointPillars} shows that our approach enjoys not only better final performance but also faster convergence speed.  
After the first fine-tuning epoch, our approach already obtains 33.06\% mAP, 8.40\% higher than training from scratch. As for SLidR, although it gets 0.93\% mAP improvement after the first epoch, its final result is inferior to that of training from scratch. This suggests that the superpixel based contrastive units are inadequate to fully drive the learning of essential semantics for downstream tasks, and its pre-training effect would be diminished when the available fine-tuning data is relatively large.

\subsection{Modality Study}
As discussed earlier, we have three choices of modalities for contrastive pre-training on point clouds: single modality (point clouds), cross-modality (images and point clouds), and multimodality (combination of the former two). 
For the single modality, we compare with the best point cloud based method ProposalContrast, as well as a stronger version (single modality +) by using our contrastive units.

\begin{table}[t]
\small
\centering
\begin{tabular}{l|ccc}
\toprule
\rowcolor[HTML]{EFEFEF}  \textbf{~Modality}  & \textbf{mAPH} & \textbf{Time} & \textbf{Memory}       \\ \hline
Scratch   &	56.59     & -      &- \\
Single Modality &	 57.59$_{+1.00}$      & 0.8$\times$     & 1.4$\times$ \\
Single Modality +  &	 58.37$_{+1.78}$      & 0.9$\times$    & 1.4$\times$ \\
Cross-Modality     &	\textbf{59.55}$_{\textbf{+2.96}}$     & 1.0$\times$    & 1.0$\times$\\
Multi-Modality     &	 58.97$_{+2.38}$         & 1.5$\times$    &1.9$\times$\\
 \bottomrule
\end{tabular}
\caption{Comparison of pre-training modalities for 3D object detection based on CenterPoint-Pillar on the validation set of WOD. We report the results of L2 mAPH, pre-training time and GPU memory.} 
\vspace{-7mm}
\label{table:waymo-modality}
\end{table}

\begin{figure}[thbp]
  \centering
  \includegraphics[width=\linewidth]{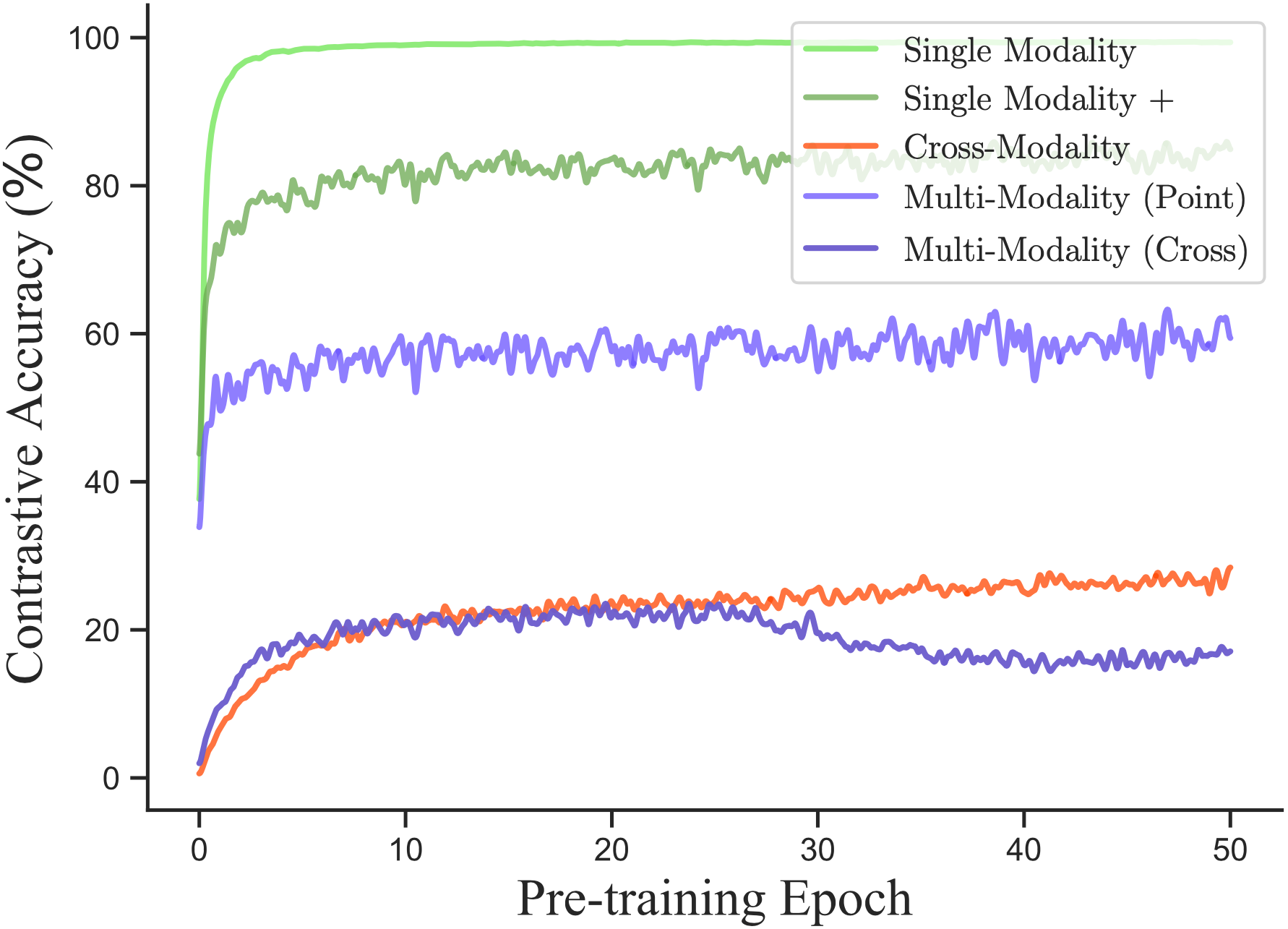}
  \vspace{-4mm}
   \caption{Comparison of the contrastive accuracy of different 
   modalities. If the similarity of a contrastive unit with its positive sample is higher than those with all negative samples, it is marked as a correct contrastive classification.} 
   \vspace{-5mm}
   \label{fig:loss_vis}
\end{figure}

As shown in Table~\ref{table:waymo-modality}, cross-modality achieves the best downstream performance with moderate pre-training time and requires the least GPU memory. We observe that pre-training on single modality tends to overfit at an early stage. Figure~\ref{fig:loss_vis} shows that the contrastive accuracy of single modality leaps to nearly 100\% after the first epoch. This indicates that point clouds provide strong hints in fitting the geometry based contrastive objective, restraining the model from learning the essential semantics. Our contrastive units help to some extent, but the overfitting (single modality +) is still obvious compared to cross-modality. As for the pre-training of multi-modality, its point cloud part or cross-modality part follows a similar trend of each individual setting, while receiving the intermediate performance, as compared in Table~\ref{table:waymo-modality}.  
Indeed, multi-modality is unnecessary since the frozen image features already act as ``anchors'', and aligning cross-modal features is a harder task. If the point cloud features of two independently augmented samples (a positive pair) are pushed close to each other, and meanwhile, they are moved toward their corresponding image features that are sufficiently close as pre-trained by 2D SSL, it is adequate to optimize the point cloud features of one sample to match to its image features.

\begin{table}[t]
\centering
\renewcommand\arraystretch{1.1}
\resizebox{0.5\textwidth}{!}{
\begin{tabular}{c|cccc}
\toprule
 \rowcolor[HTML]{EFEFEF}  \textbf{~Pre-training} & \textbf{1\%} & \textbf{5\%} & \textbf{10\%} & \textbf{20\%}       \\ \hline

Scratch$^*$    &  -  &	44.35     & 51.14      &55.25 \\

PointContrast$^*$~\cite{xie2020pointcontrast}    &  -  &	44.97$_{+0.62}$     & 52.35$_{+1.21}$     &56.31$_{+1.06}$\\

GCC-3D$^*$~\cite{Liang_2021_ICCV}&  -  &	 47.85$_{+3.50}$      & 53.89$_{+2.75}$     &57.39$_{+2.14}$\\
\hline

Scratch    &  26.05  &	47.17      & 52.73     &56.59\\
SLidR~\cite{SLidR}       &    31.03$_{+4.98}$ &  	49.90$_{+2.73}$   &    54.46$_{+1.73}$  & 57.23$_{+0.64}$ \\
ProposalContrast~\cite{yin2022proposalcontrast}       &    33.30$_{+7.25}$ &  	51.60$_{+4.43}$   &   55.67$_{+2.94}$   &57.59$_{+1.00}$ \\
Ours        & \textbf{ 38.55$_{\textbf{+12.50}}$}   & \textbf{ 54.62$_{\textbf{+7.45}}$}	   &    \textbf{57.35$_{\textbf{+4.62}}$}  & \textbf{59.55$_{\textbf{+2.96}}$} \\ \bottomrule
\end{tabular}}
\caption{Comparison of 3D object detection based on CenterPoint-Pillar with different fractions of data. We report the results of L2 mAPH on the validation set of WOD. ${ }^*$ denotes the results from~\cite{Liang_2021_ICCV}.}
\vspace{-1.5mm}
\label{table:waymo-CenterPoint-PointPillars-data-efficient}
\end{table}

\begin{table}[t]
\centering
\resizebox{\linewidth}{!}{
\begin{tabular}{c|cccc}
\toprule
\rowcolor[HTML]{EFEFEF}  \textbf{Method}&   {Scratch} & {SLidR~\cite{SLidR}}             & {ProposalContrast~\cite{yin2022proposalcontrast}}    &    {Ours}       \\ \hline 
mAP  &	 49.40  &  49.87	&	50.87 & \textbf{52.20}  \\ \bottomrule 
\end{tabular}}
\vspace{-1mm}
\caption{Comparison of 3D object detection based on CenterPoint-Pillar under the transfer learning setting. We report the results of mAP on the validation set of ONCE.}
\vspace{-5mm}
\label{table:once-pillar}
\end{table}

\begin{table}[h]
\small
\centering
\renewcommand\arraystretch{1.1}
\resizebox{0.5\textwidth}{!}{
\begin{tabular}{c|cc|cc}
\toprule
\rowcolor[HTML]{EFEFEF}  \textbf{~Pre-training} & \textbf{nuScenes}             & \textbf{Gain} & \textbf{SemanticKITTI}            & \textbf{Gain}         \\ \hline
Scratch    &  30.3 &	-   &  39.5 & - \\
PointContrast~\cite{xie2020pointcontrast} & 32.5 & 2.2   &  41.1 & 1.6 \\
DepthContrast~\cite{Zhang_2021_ICCV} &  31.7 & 	1.4   &41.5   & 2.0 \\
PPKT~\cite{liu2021ppkt}  &  37.8 & 	7.5   & 43.9  & 4.4 \\
SLidR~\cite{SLidR}       &    38.3 &  	8.0    &   44.6&  5.1 \\
SegContrast~\cite{segcontrast}       &    31.9 &  	1.6    &   - &  -\\

Ours        & \textbf{39.2}   & \textbf{8.9}	  & \textbf{45.7 }&\textbf{6.2} \\ \bottomrule
\end{tabular}}
\vspace{-1mm}
\caption{Comparison of 3D semantic segmentation based on MinkowskiNet. We report the results of mIOU on the validation sets of nuScenes and SemanticKITTI.} 
\label{table:nuscenes-seg-efficient}
\vspace{-1mm}
\end{table}

We further quantitatively compare the feature alignment under cross-modality and multi-modality. Specifically, we compute the feature cosine similarity of a positive pair as the alignment score. 
By randomly sampling $1\times10^{7}$ positive pairs, we observe that the cross-modal pre-training has a much higher alignment score (0.708) than that (0.532) of the multi-modal pre-training. This again shows the advantage of cross-modality over multi-modality in contrastive learning of point clouds.

\subsection{Data-Efficient Fine-Tuning and Transfer Learning}
\label{sec:Data Efficient Fine-tuning}

\noindent\textbf{Comparison on 3D Object Detection.} 
We gradually increase the amount of annotated training data from 1\%, 5\%, 10\%, to 20\%, and evaluate the fine-tuning performance on WOD. 
As shown in Table~\ref{table:waymo-CenterPoint-PointPillars-data-efficient}, our approach exhibits greater performance gains over other methods when a small quantity of labeled data is available. For instance, when merely having 1\% data, we observe a dramatic improvement of 12.50\% mAPH, which substantially outperforms other methods. 
It is also interesting to note that with 10\% data, we beat the performance of training from scratch using the standard setting of 20\% data, meaning that human labeling efforts can be halved with our approach.   

We next study the transfer learning capability of our approach. Specifically, we adopt CenterPoint-Pillar pre-trained on WOD, then fine-tune and evaluate on the standard training and validation sets of ONCE. As shown in Table~\ref{table:once-pillar}, our approach achieves superior improvement (+2.80\% mAP) over training from scratch, which largely outperforms SLidR (+0.47\% mAP) and ProposalContrast (+1.47\% mAP), suggesting the strong generalizability of our approach.

\noindent\textbf{Comparison on 3D Semantic Segmentation.} 
We extend our approach to 3D semantic segmentation, where we first pre-train MinkowskiNet on nuScenes, and then fine-tune on nuScenes as well as SemanticKITTI (transfer learning evaluation), following the experimental settings in SLidR. As compared in Table~\ref{table:nuscenes-seg-efficient}, with 1\% of labeled data, our approach achieve 8.9\% and 6.2\% performance gains on nuScenes and SemanticKITTI, exceeding the improvement by other methods. For this downstream task, the point cloud only based pre-training methods (PointContrast and DepthContrast) produce much lower improvement in comparison to the cross-modal pre-training methods (PPKT, SLidR, and ours), which indicates the benefit of visual information provided by image features in contrastive pre-training to facilitate this fine-grained point-wise perception task.

\begin{table}[htbp]
\vspace{-0.1em}
\centering
\resizebox{0.95\linewidth}{!}{
\begin{tabular}{ccccc|cc}
\toprule
\rowcolor[HTML]{EFEFEF} \textbf{Pillar} & \textbf{Neighbor} & \textbf{Instance} & \textbf{Similarity} & \textbf{FPS} & \textbf{mAPH} \\ \hline
\checkmark & & \checkmark & & \checkmark & 37.99 \\ 
\checkmark & & \checkmark & \checkmark & \checkmark & 38.34  \\ \hline
& \checkmark & & & \checkmark & 36.51 \\
& \checkmark & \checkmark & & \checkmark & 37.86  \\
& \checkmark & & \checkmark & \checkmark & 37.01 \\
& \checkmark & \checkmark & \checkmark &  & 37.02  \\
& \checkmark & \checkmark & \checkmark & \checkmark & \textbf{38.55} \\ \bottomrule 
\end{tabular}}
\caption{Ablation study of the different combinations of feature aggregation based on nearest neighbor and pillar, instance-aware clustering, similarity-balanced sampling and FPS. We report L2 mAPH using CenterPoint-Pillar on the validation set of WOD under the data-efficient setting.} 
\vspace{-1em}
\label{tab:ablation}
\end{table}

\subsection{Ablation Study}
\noindent\textbf{Each Individual Component. }
Here we perform various ablation experiments under the 1\% data-efficient setting to understand each individual component of our approach, as shown in Table~\ref{tab:ablation}. We first evaluate different ways of feature aggregation for a contrastive unit as mentioned in Section~\ref{sec:initial}, including $K$ nearest neighboring points of the unit or the points within the pillar centered at the unit. It is observed that the two ways are overall comparable, showing the flexibility of our approach. Compared to the full pre-training framework, removing either instance-aware clustering or similarity-balanced sampling results in a performance drop. Moreover, changing farthest point sampling to random sampling for initial contrastive units leads to lower performance. These ablation study results collectively validate the proposed contrastive unit design.

\noindent\textbf{Image Backbone.}
By default, we adopt MoCoV3~\cite{chen2021empirical}, which is a CNN (ResNet50) based image backbone. We further experiment with the Transformers (Swin-T) based image backbone EsViT~\cite{li2021esvit}. We extract and concatenate image features from three levels, and conduct the same contrastive pre-training for 20 epochs. As shown in Table~\ref{table:ablation-image-backbone}, EsViT also brings obvious improvement compared to training from scratch, but its improvement is inferior to that of MoCoV3. We suspect that the patch encoding in Transformers partially breaks the point-pixel feature correspondence because of the large patch size. 
Moreover, we apply a randomly initialized ResNet50 to extract image features. As shown in Table~\ref{table:ablation-image-backbone}, the randomly initialized image backbone only slightly improves the performance compared with training from scratch. This reveals that the SSL pre-trained image network is crucial to the success of cross-modal contrastive pre-training.

\begin{table}[h]
\centering
\resizebox{0.5\textwidth}{!}{
\begin{tabular}{c|cccc}
\toprule
\rowcolor[HTML]{EFEFEF}  \textbf{Image Backbone}&   {Scratch} & {MoCoV3}             & {EsViT}    &    {Random}       \\ \hline 
mAP  &	 60.74 &  62.89 	&	62.51 & 61.18  \\ 
mAPH  &	 56.59 &  59.02  &	58.54   & 56.83 \\
\bottomrule 
\end{tabular}}
\vspace{-1mm}
\caption{Comparison of different image backbones used in our cross-modal contrastive learning.}
\vspace{-1mm}
\label{table:ablation-image-backbone}
\end{table}

\noindent\textbf{Image Feature Levels.}
Given the frozen image backbone MoCoV3, we extract and combine image features from different levels. Based on the three levels corresponding to the scales of 1/4~(P2), 1/8~(P3), and 1/16~(P4) of the input image size, we evaluate the three design choices, namely P2, P4, and P2+P3+P4 after pretraining for 20 epochs. As shown in Table~\ref{table:ablation:image-feature-level}, the image features concatenated from three levels achieve the best performance, and P4 outperforms P2 due to its high-level semantics from deep abstraction. This suggests that image features from different levels are more advantageous to cross-modal contrastive learning. 

\begin{table}[h]
\small
\centering
\resizebox{0.43\textwidth}{!}{
\begin{tabular}{c|ccc}
\toprule
\rowcolor[HTML]{EFEFEF}  \textbf{Image Feature Level}         & {P2}   & {P4}       & {P2+P3+P4}     \\ \hline
mAP    &	62.09  &	62.67  &  62.89  \\
mAPH    &	58.01  &	58.69  &  59.02 \\
\bottomrule
\end{tabular}}
\vspace{-1mm}
\caption{Comparison of different image feature levels used in our cross-modal contrastive learning.}
\label{table:ablation:image-feature-level}
\vspace{-5mm}
\end{table}

\section{Related Work}

\noindent\textbf{Self-Supervised Learning in 2D.}  
Early works hinging on pretext tasks~\cite{gidaris2018rotation, zhang2016colorful} are limited to learning low-level cues. More recent contrastive learning methods like MoCo~\cite{byol} and SimCLR~\cite{pmlr-v119-chen20j} align the features of augmentations from the same image while pushing away other images, and achieve similar linear probing performance to fully supervised pre-training. Masked image modeling (MIM)~\cite{beit} employs a high mask ratio to reconstruct an image in a generative way and shows promising results.

\noindent\textbf{Self-Supervised Learning in 3D.} 
Inspired by 2D SSL, contrastive learning and masked modeling are the two main tracks for 3D SSL. PointContrast~\cite{xie2020pointcontrast} exploits point clouds from two views to build the contrastive pre-training objective. DepthContrast~\cite{Zhang_2021_ICCV} applies both point and voxel based backbones to extract features of each contrastive unit.  
Recently, GCC-3D~\cite{Liang_2021_ICCV} introduces a two-stage pre-training paradigm to treat a local neighborhood and motion group as the contrastive unit. In~\cite{yin2022proposalcontrast}, ProposalContrast uses proposals and online clustering to perform contrastive pre-training. 
Another line is to conduct contrastive learning across images and point clouds to alleviate the limitations of geometry-only SSL. Pri3D~\cite{hou2021pri3d} and PPKT~\cite{liu2021ppkt} use pixel-point correspondences to build the pre-training objective. 
SLidR~\cite{SLidR} introduces superpixels to group neighboring pixels in 2D 
as a region-level contrastive unit. 
Generative masked auto-encoding has also been explored for point clouds. Occupancy-MAE~\cite{min2022occupancymae}, BEV-MAE~\cite{lin2022bevmae} and MAELi~\cite{krispel2022maeli} all utilize a masked auto-encoder in the space of BEV. 
GeoMAE~\cite{tian2023geomae}  further utilizes geometry clues such as surface normal and curvature as the self-supervised objective.

\section{Conclusion}

We present a cross-modal self-supervised learning framework with the proposed effective contrastive units for self-driving point clouds. We provide the first comprehensive modality study of contrastive learning for LiDAR, and show that cross-modal learning performs the best for both pre-training efficiency and downstream improvement. Our contrastive units facilitate contrastive pre-training via the design of instance-aware clustering and similarity-balanced sampling. Extensive experiments reveal that our approach achieves remarkable performance gains. We hope our findings would encourage the research community on the cross-modal and more targeted designs of self-driving point clouds.   







\addtolength{\textheight}{-12cm}   

\bibliographystyle{IEEEtran}
\balance
\bibliography{main}

\end{document}